%% file: ICRA19.tex
\documentclass[letterpaper, 10pt, conference]{ieeeconf}

\pdfoutput=1
\IEEEoverridecommandlockouts                              
\usepackage[utf8]{inputenc}
\usepackage[T1]{fontenc}

\usepackage[pdftex]{graphicx}
\usepackage{epstopdf}
\usepackage{amsmath}
\usepackage{amssymb}
\usepackage{nicefrac}
\usepackage{booktabs}
\usepackage{multirow}
\usepackage{siunitx}
\usepackage{microtype}
\usepackage[vlined]{algorithm2e}
\usepackage[table,xcdraw]{xcolor}
\usepackage{multicol}
\usepackage{flushend}
\usepackage{fancyhdr}

\makeatletter
\let\NAT@parse\undefined
\makeatother
\usepackage[numbers,sort,square]{natbib}

\usepackage{hyperref}

\def\equationautorefname~#1\null{(#1)\null}

\graphicspath{{./img/}}

\newcommand{\bestResult}[1]{{\color[HTML]{32CB00} \textbf{#1}}}

\input{math.tex}

\title{\LARGE \bf Expectation-Maximization for Adaptive Mixture Models in Graph Optimization}

\author{\authorblockN{Tim Pfeifer and Peter Protzel}%
\authorblockA{Dept.~of Electrical Engineering and Information Technology\\
TU Chemnitz, Germany\\
Email: \{firstname.lastname\}@etit.tu-chemnitz.de}%
\thanks{The project is funded by the "Bundesministerium f\"ur Wirtschaft und Energie" (German Federal Ministry for Economic Affairs and Energy).}%
}

\begin{document}


\maketitle
\thispagestyle{fancy}
\fancyhf{}
\fancyhead[OL]{ 
	\footnotesize
	Proc. of IEEE International Conference on Robotics and Automation (ICRA), 2019, Montreal, Canada. DOI: 10.1109/ICRA.2019.8793601\\
	\tiny
	\copyright 2019 IEEE. Personal use of this material is permitted. Permission from IEEE must be obtained for all other uses, in any current or future media, including
	reprinting/republishing this material for advertising or promotional purposes, creating new collective works, for resale or redistribution to servers or lists, or reuse of any copyrighted component of this work in other works.}

\addtolength{\headheight}{\baselineskip}

\begin{abstract}
Non-Gaussian and multimodal distributions are an important part of many recent robust sensor fusion algorithms.
In difference to robust cost functions, they are probabilistically founded and have good convergence properties.
Since their robustness depends on a close approximation of the real error distribution, their parametrization is crucial.

We propose a novel approach that allows to adapt a multi-modal Gaussian mixture model to the error distribution of a sensor fusion problem.
By combining expectation-maximization and non-linear least squares optimization, we are able to provide a computationally efficient solution with well-behaved convergence properties.

We demonstrate the performance of these algorithms on several real-world GNSS and indoor localization datasets.
The proposed adaptive mixture algorithm outperforms state-of-the-art approaches with static parametrization.
Source code and datasets are available under \url{https://mytuc.org/libRSF}.
\end{abstract}

\section{Introduction}

Robotic systems as well as autonomous vehicles require a reliable estimation of their current state and location.
The algorithms that compute this information from sensor data are typically formulated as least squares optimization and solved with frameworks like Ceres \cite{Agarwal} or GTSAM \cite{Dellaert}.
Usually, they rely on the assumption of Gaussian distributed measurement errors, which is correct for many sensors and comes with advantageous mathematical consequences for the estimation process.
However, there is a broad range of sensors, like global navigation satellite systems (GNSS), wireless range measurements, ultrasonic range finders or vision-based systems, that violate this assumption.
Even simple wheel odometry can slip on difficult grounds and cause non-Gaussian error distributions.
These violations can heavily distort the estimation process and lead to false estimates.
A variety of robust approaches exist to provide robustness against non-Gaussian errors, but many of them require an exact parametrization and therefore knowledge about the expected error distribution.
Previous evaluations \cite{Latif2014, Pfeifer2016} have shown that the optimal set of parameters is hard to find and small deviations can lead to fatal errors.

\begin{figure}
\centering
\includegraphics[width=\linewidth]{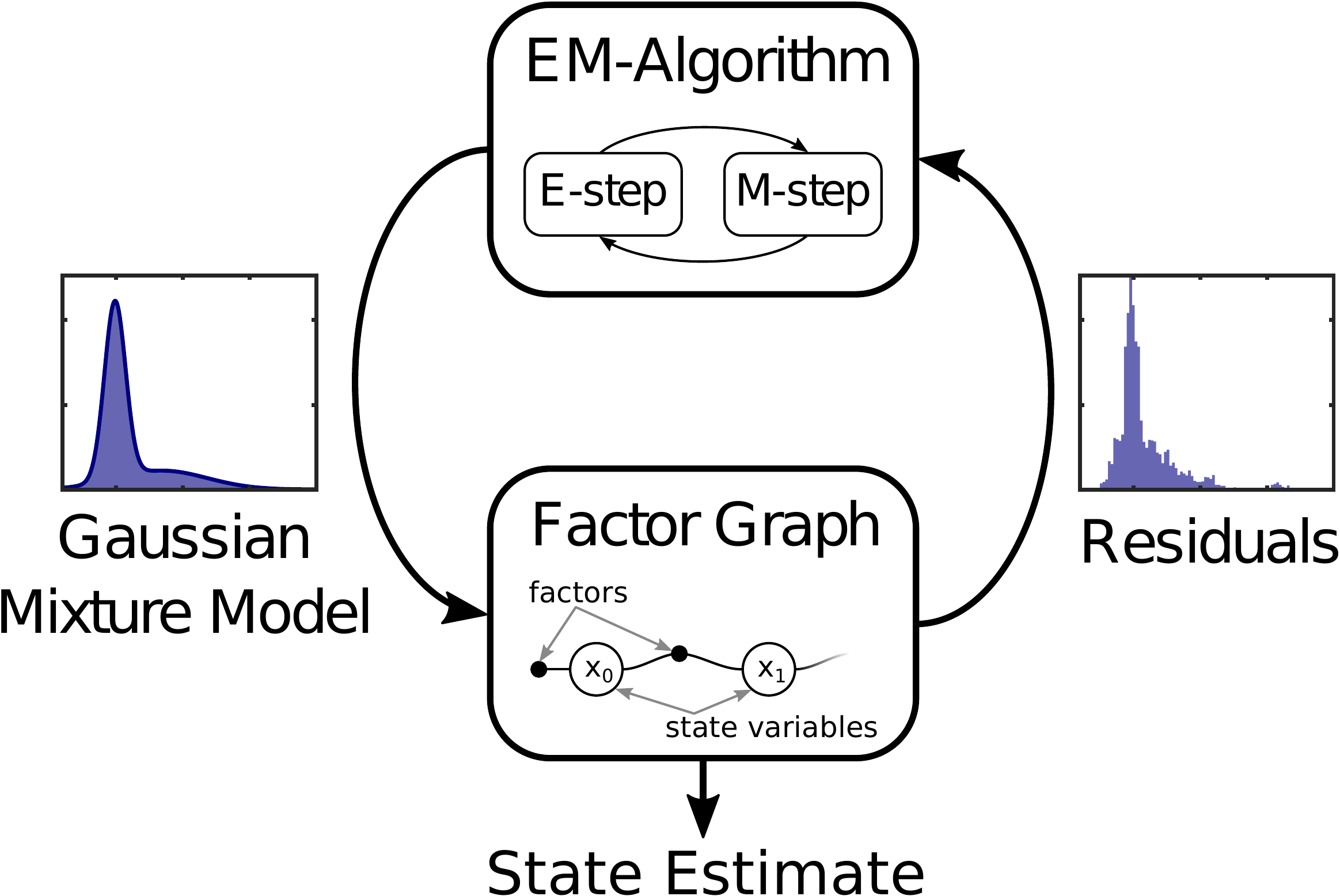}
\caption{The main concept of our proposed adaptive mixtures algorithm. Residuals of the optimization problem (represented as factor graph) are used to estimate a multimodal error model. Expectation-Maximization (EM) is applied for a well-behaved convergence.}
\label{fig:Concept}
\end{figure}

With this paper, we want to introduce a novel approach to estimate the sensors' error distribution during the least squares optimization process.
As improvement of the ideas in \cite{Pfeifer2018}, we show how to construct an expectation-maximization (EM) based algorithm that adapts a Gaussian mixture model (GMM) to the sensor properties.
\autoref{fig:Concept} shows how the least squares optimization and the error model estimation are connected.
Due to its graphical representation, the least squares problem is also referred to as factor graph.
In difference to previous work, we are able to overcome the limitations of the Max-Mixture approximation and implement an exact GMM inside the least squares problem.
We also increase the robustness of the GMM's estimation process without extensive parametrization.
This results in a sensor fusion algorithm that is robust against non-Gaussian errors without prior knowledge.
Through extensive evaluation with real world localization datasets from \cite{Suenderhauf2013a, Reisdorf2016, Pfeifer2017}, we are able to demonstrate its performance in comparison to several state-of-the-art approaches.
Our datasets represent different scenarios from GNSS localization in urban canyons to centimeter-level wireless ranging.


\section{Prior Work}\label{sec:prior}

Motivated by the simultaneous localization and mapping (SLAM) problem, several algorithms exist that try to achieve robustness against non-Gaussian outliers.
We want to give a brief overview of them and show why existing solutions are often difficult to apply to real world problems.

Several methods use generic mechanisms to reduce the influence of outliers without knowledge about the true sensor distribution.
A typical example are M-estimators, that apply non-convex cost functions to the residual term of the optimization problem.
They are suited for many problems, but the right parametrization is hard to find and global convergence is not guaranteed.
A different approach is Switchable Constraints (SC) \cite{Suenderhauf2012a} that introduced a set of additional weights, each assigned to one measurement.
The weights are optimized to weight possible outliers down, which works on SLAM \cite{Suenderhauf2013b} and GNSS localization problems \cite{Suenderhauf2013a}.
However, the introduced tuning parameter is not directly connected to probabilistic metrics and has to be set manually.
This can be very difficult as shown in \cite{Latif2014} and \cite{Pfeifer2016}.
Dynamic Covariance Scaling (DCS) \cite{Agarwal2013} was introduced as improvement of SC and is basically an M-estimator.
In difference to SC, the weights are not longer part of the incremental optimization process.
Instead, they get optimized analytical, which leads to an overall faster convergence.
However, they keep the disadvantage of difficult parametrization \cite{Latif2014, Pfeifer2017}.
Although there is a usable parameter window for many SLAM applications, this does not necessarily apply to general sensor fusion.
To overcome the parametrization of SC/DCS for non-SLAM applications, we introduced Dynamic Covariance Estimation (DCE) in \cite{Pfeifer2017}.
Due to the non-convex optimization surface, this approach is still limited to problems with a good initialization and moderate outliers.
\citeauthor{Agamennoni2015} proposed an approach to tune the parameters of some M-estimators in \cite{Agamennoni2015}.
It is limited to a subset of M-estimators that can be described as elliptical distributions and cannot be applied to more robust ones like DCS or the Tukey M-estimator.

Another approach to handle non-Gaussian measurements, is to consider their true distribution during the estimation process.
In difference to previously mentioned methods, this allows to handle asymmetric or multimodal distributions probabilistically correct.
Max-Mixture (MM) \cite{Olson2012} describes the expected distribution with an approximation of a GMM.
\citeauthor{Rosen2013} introduced a method that allows arbitrary non-Gaussian distributions \cite{Rosen2013}.
A drawback of both approaches is the lack of concepts to get the required parametrization of the GMM or any other non-Gaussian distribution.
Existing approaches \cite{Morton2013, Rosen2013a} estimate it in advance which is not possible if the error distribution depends on the environment and varies over time.
This paper is build on top of both ideas from \citeauthor{Olson2012} \cite{Olson2012} / \citeauthor{Rosen2013} \cite{Rosen2013}, so we provide more details about their concepts in \autoref{sec:gmm}.
A recently published algorithm \cite{Fourie2016} allows interference of non-parametric distributions based on kernel densities.
This offers new possibilities for non-Gaussian distributions but cannot solve the parametrization problem.

Our recently published approach of self-tuning GMMs \cite{Pfeifer2018} aims to overcome the burden of parametrization by introducing a self-optimizing version of Olson's Max-Mixture.
By enabling the optimizer to change the model's parameters, the GMM can be adapted to the most likely error distribution.
Although the convergence on a real world GNSS dataset is shown, the direct optimization of the error model comes with several drawbacks:
\begin{enumerate}
	\item The algorithm requires a good initial guess of the true distribution since it can not recover from a local minimum.
	
	\item Tight bounds for the optimized GMM parameters are required to work around numerical issues.
	
	\item The applied Max-Mixture model is just an approximation of a GMM which is not correct for GMMs with strongly overlapping components.
	
	\item The used model is fixed in the number of mixture components and there is no possibility to detect if more or fewer components are required.
\end{enumerate}
Therefore, \cite{Pfeifer2018} is just a proof-of-concept and real applications require further steps.
With this paper we relax limitation (1) and overcome (2) and (3) completely.
We want to introduce a sensor fusion algorithm that adapts its error model during runtime without more than minimal knowledge about the true distribution.
As far as we know, this is the first algorithm which implements a robust self-optimizing GMM as error model of a realtime capable state estimation algorithm.

\section{Gaussian Mixture Models}\label{sec:gmm}

For the representation of multimodal distributions, Gaussian mixture models are a state-of-the-art approach with several advantages.
With a weighted sum of multiple Gaussians, corresponding to \autoref{eqn:gaussian_sum}, non-Gaussian properties like asymmetry or multiple modes can be represented easily.
Due to the variable number of components, distributions with different complexity can be described.
With expectation-maximization exists a powerful algorithm to estimate the model's parameters from distributed data.
We want to show in this section how GMMs can be used for least squares.
Although this is nothing new, it is important for the further construction of the proposed algorithm.
\begin{equation}
\prob \sim \sum_{j} \weight_{j} \cdot \mathcal{N}(\mean_j,\cov_j) 
\label{eqn:gaussian_sum}
\end{equation}

\subsection{Multivariate Gaussians and Least Squares}
The least squares problem, which is solved for state estimation, arises from the formulation of a maximum-likelihood problem \autoref{eqn:argmax} where $\Measurement$ is a set of measurements $\measurementi$ and $\State$ is the set of estimated states $\statei$.
\begin{equation}
\StateOpt =\argmax_{\State} \probOf{\State}{\Measurement}
\label{eqn:argmax}
\end{equation}
To estimate the optimal set of states $\StateOpt$, the problem can be described as product of conditional probabilities:
\begin{equation}
\probOf{\State}{\Measurement} \propto \prod_i \probOf{\measurementi}{\statei}
\label{eqn:posteriori}
\end{equation}
By applying the negative logarithm, the optimization problem can be rewritten as sum:
\begin{equation}
\StateEst = \argmin_{\State} \sum_{i} -\ln(\probOf{\measurementi}{\statei})
\label{eqn:log_like}
\end{equation}
The estimated $\StateEst$ is the maximum-likelihood estimator of $\State$.

For multivariate Gaussians, the corresponding least squares problem is defined as:
\begin{equation}
\StateEst = \argmin_{\State} \sum_{i} \frac{1}{2} \left\| \sqrtinfo \left( \errori - \mean \right) \right\| ^2
\label{eqn:arg_min}
\end{equation}
The estimation error $\error(\statei,\measurementi)$ is defined as non-linear function of a measurement and a corresponding subset of states.
Instead of the covariance Matrix $\cov$, we use the square root information matrix $\sqrtinfo$ that is defined by $\sqrtinfo = \cov^{-\frac{1}{2}}$.
It can be computed from ${\cov}^{-1}$ using Cholesky decomposition.
For a sum of Gaussians with $n$ components, the conditional probability is defined as:
\begin{equation}
\begin{split}
\prob(\measurementi|\statei) \propto \sum_{j=1}^n&{c_j \cdot \exp \left( -\frac{1}{2} \left\| \sqrtinfoj \left( \errori - \mean_j \right) \right\| ^2 \right)}\\
\text{with } &c_j = \weight_j \cdot \det\left( \sqrtinfoj \right)
\end{split}
\label{eqn:gmm}
\end{equation}
Due to the summation, the logarithm cannot be pushed inside and the log-likelihood has to be calculated differently.
\citeauthor{Olson2012} \cite{Olson2012} respectively \citeauthor{Rosen2013} \cite{Rosen2013} provided two possible solutions.
	
\subsection{Approximated Solution (Max-Mixture)}\label{ssec:mm}

In \cite{Olson2012} the summation of a GMM was replaced by the maximum operator, which leads to:
\begin{equation}
-\ln(\boldsymbol{P}) 
= -\ln 
	\max_{j}
	\left(
		 c_j \cdot \exp \left( -\frac{1}{2} \left\| \sqrtinfoj \left( \errori - \mean_j \right) \right\| ^2 \right)
	\right)
\label{eqn:loglike_mm1}
\end{equation}
This approximation is valid as long as the Gaussian components are well separated.
The maximum becomes a minimum when the logarithm is pushed inside:
\begin{equation}
-\ln(\boldsymbol{P}) 
= 
\min_{j}
\left(
- \ln c_j + 
\frac{1}{2} \left\| \sqrtinfoj \left( \errori - \mean_j \right) \right\| ^2 
 \right)
\label{eqn:loglike_mm2}
\end{equation}
In difference to other implementations, we keep the log-normalization term $- \ln c_j$ to preserve a consistent optimization surface.
Hence, we introduce a normalization constant $\gamma_m$ to construct a well-behaved least squares problem:
\begin{equation}
\begin{split}
\StateEst = 
\argmin_{\State} 
&\sum_{i} 
\frac{1}{2}
\min_{j}
\begin{Vmatrix}
\sqrt{- 2 \cdot \ln \frac{c_j}{\gamma_m}}\\
\sqrtinfoj \left( \errori - \mean_j \right)
\end{Vmatrix}
^2\\
&\text{with } \gamma_m = \max_{j} c_j
\end{split}
\label{eqn:arg_min_mm}
\end{equation}
The additional term $- \ln \frac{c_j}{\gamma_m}$ is treated as separate dimension of the vectorized error function.
A positive expression under the square root is guaranteed by $\gamma_m$.
The separation of log-normalization and quadratic error leads to better convergence, since the partial derivative regarding the state value $\statei$ is identical to the original problem \autoref{eqn:arg_min}.

\subsection{Exact Solution (Sum-Mixture)}\label{ssec:sm}

The exact implementation of a GMM inside a least squares problem is possible with the approach proposed in \cite{Rosen2013}.
\citeauthor{Rosen2013} demonstrated that arbitrary distributions can be applied to weight the estimation error with:
\begin{equation}
\begin{split}
\StateEst = 
\argmin_{\State} 
&\sum_{i} 
\left\| \sqrt{
	-\ln   
	 \left(
	 \frac{\probOf{\measurementi}{\statei}}{\gamma_s}
	 \right) 
}\right\| ^2\\
&\text{with } \gamma_s \geq \max_i \left(\probOf{\measurementi}{\statei} \right)
\end{split}
\label{eqn:arg_min_sm}
\end{equation}
It requires a normalization constant $\gamma_s$ to keep the negative log-likelihood positive.
Based on \autoref{eqn:gmm} the new normalized probability of a GMM is defined as:
\begin{equation}
\frac{\probOf{\measurementi}{\statei}}{\gamma_s}
= 
\sum_{j}{ \frac{c_j}{\gamma_s} \cdot \exp \left( -\frac{1}{2} \left\| \sqrtinfoj \left( \errori - \mean_j \right) \right\| ^2 \right)}
\label{eqn:loglike_sm}
\end{equation}
Since it has to satisfy $\nicefrac{\probOf{\measurementi}{\statei}}{\gamma_s} \leq 1$, the normalization can be defined as summation of all constant terms $\gamma_s = \sum_{j} c_j$.
The resulting lest squares problem is:
\begin{equation}
\StateEst = \argmin_{\State} \sum_{i} \left\| 
\sqrt{-\ln   
		\left(  
		\sum_{j}{ \frac{c_j}{\gamma_s} \cdot \exp \left(
																\dots
													   \right)}
		\right) 
	}\right\| ^2
\label{eqn:arg_min_sm2}
\end{equation}

\section{Expectation-Maximization for Mixture Distributions}\label{sec:em}

The EM-algorithm \cite{Dempster1977} is a maximum likelihood approach in which a set of parameters $\Params$ is estimated from a set of observed variables $\observedvar$, but also depends on a set of hidden ones $\hiddenvar$.
Since $\Params$ and $\hiddenvar$ are unknown at the beginning, $\Params$ cannot be estimated directly.
Instead, an alternating sequence of E-steps and M-steps is performed.
E-steps estimate the hidden variables based on an initial guess of $\Params$ and M-steps compute $\Params$ based on the previously estimated $\hiddenvar$.
Both steps are performed iteratively until a maximum number of iterations or a different convergence criterion is reached. 
For GMM estimation it is defined as:
\begin{equation}
\begin{split}
	&\observedvar^{GMM} = \left\lbrace  \errori \right\rbrace \hspace{0.5cm}
	\hiddenvar^{GMM} = \left\lbrace \alpha_{ij}\right\rbrace \\
	&\hspace{0.8cm} \Params^{GMM} = \left\lbrace \weight_j, \mean_j, \sqrtinfoj \right\rbrace
\end{split}
\end{equation}
The observed variable is the measurement error $\errori = \measurementi - f(\statei)$, defined by the measured value $\measurementi$ and the true value of the state $\statei$.
Hidden variable is the probability $\alpha_{ij}$ of each measurement $\measurementi$ to belong to component $j$ of the GMM.
The parameters $\weight_j, \mean_j$ and $\sqrtinfoj$ define the mixture distribution.
For mathematical details please see Appendix \ref{Appendix_EM}.

\section{Self-Tuning Mixtures}\label{sec:adaptive}

To solve the problem of simultaneous state and error model estimation, we define them together as a nested EM-algorithm.
At first, we assume that we are able to estimate a state that is a coarse approximation of the true state $\State$.
This can be done with a simple initialization using accumulated odometry measurements or non-robust least squares optimization.
It implies that the estimated measurement error $\erroriEst$ is also close to it's true value:
\begin{equation}
 \stateiEst \approx \statei \Rightarrow \erroriEst \approx \errori
\end{equation}
Based on this assumption, we can redefine the observed variable for the GMM estimation problem as \autoref{eqn:observed_gmm} and the EM problem of state estimation as \autoref{eqn:em-se}.

\begin{equation}
\observedvar^{GMM} = \left\lbrace  \erroriEst = \error(\stateiEst, \measurementi) \right\rbrace
\label{eqn:observed_gmm}
\end{equation}
\begin{equation}
	\observedvar^{SE} = \Measurement \hspace{0.5cm}
	\hiddenvar^{SE} = \Params^{GMM} \hspace{0.5cm}
	\Params^{SE} = \State 
	\label{eqn:em-se}
\end{equation}

Algorithm \ref{algo:adaptive} summarizes how both stages of EM are connected.
After the initialization of $\State$ and $\Params^{GMM}$, an outer EM-algorithm for state estimation (SE) is combined with an inner one for the Gaussian mixture model (GMM).
The E-step of the SE problem estimates the hidden parameter $\Params^{GMM}$ and is composed of the E- and the M-step of the GMM estimation problem.
The outer M-step is the least squares optimization according to the Max-Mixture \autoref{eqn:arg_min_mm} or Sum-Mixture \autoref{eqn:arg_min_sm2} formulation from \autoref{sec:gmm}.
To keep the algorithm real-time capable, only the measurements and states in a sliding window with the length $t_{SW}$ are considered in the estimation problem.

\begin{algorithm}[htbp]
	\SetAlgoLined
	
	\KwResult{$\StateEst, \ParamsEst$}
	
	Initialize states $\State$ randomly
	
	Estimate $\State_{t=0}$ with \autoref{eqn:arg_min}

	Initialize $\Params^{GMM}_{t=0}$ (e.g with \autoref{eqn:init_gmm})

	\ForEach{time step $t$}
	{
		Add $\measurementi$, $\statei$ at $t$ to $\prob(\State|\Measurement, \Params^{GMM})$
		
		Remove $\measurementi$, $\statei$ older than $t-t_{SW}$ from $\prob(\State|\Measurement, \Params^{GMM})$
		
		\tcp{E-step state estimation}
		\Begin(estimate $\Params^{GMM}_{t}$) 
		{
			Compute all $\errori$ from $\State, \Measurement$
			
			Initialize $\Params^{GMM}_{t} \leftarrow \Params^{GMM}_{t=0}$
			
			\Repeat{convergence}
			{
				\tcp{E-step GMM}
				Compute $\alpha_{ij}$ from $\errori, \Params^{GMM}_{t}$ with \autoref{eqn:e-step}
				
				\tcp{M-step GMM}
				Compute $\Params^{GMM}_t$ from $\errori, \alpha_{ij}$ with \autoref{eqn:m-step}
			}
		}
		
		Update $\prob(\State|\Measurement, \Params^{GMM})$ with $\Params^{GMM}_t$
		
		\tcp{M-step state estimation}
		Estimate $\State_t$ with \autoref{eqn:arg_min_mm} or \autoref{eqn:arg_min_sm2}

	}
	\caption{Nested EM for Self-tuning Mixtures}
	\label{algo:adaptive}
\end{algorithm}

Since the outer EM-algorithm is computed only once per time step, its convergence have to be achieved over time.
For sensor fusion applications, the difference between each time step is small, so it is called often enough to converge until the distribution parameters change.
The convergence of the algorithm is hard to prove, since it depends on the overall structure of the estimation problem and the non-linearity in the error function $\error$.
It has to be assumed that the optimized surface of the problem has several local minima, which is a common problem in non-linear state estimation.
Nevertheless, we would like to mention a few arguments in favor of convergence:
\begin{enumerate}
	\item The resulting log-likelihood of Max-Mixture and Sum-Mixture is mostly convex which supports convergence.
	
	\item The inner EM-algorithm for the GMM will converge for sure. \cite[p. 450]{Bishop2006} 
	
	\item Since the algorithm is solved iteratively, every time step starts close to the minima that the previous one found.	
\end{enumerate}
The crucial question is, if a good initial starting point for the state values can be found without knowing the exact distribution.
As we will show in our experimental evaluation, this is possible at least for different localization problems.

\section{Localization Problems as Factor Graph}\label{sec:evaluation}

Since we evaluate the proposed adaptive GMM approach on different localization datasets, we want to explain what makes them challenging and how the estimation problems are composed.
Our adaptive mixtures algorithm is applied to reduce the estimation error that non-line-of-sight (NLOS) measurements cause.
Hence, the estimation error $\errori$, that is described with the Gaussian mixture, is related to the one-dimensional (pseudo) range measurements.

\subsection{UWB-Radio Localization}

In \cite{Pfeifer2017}, we introduced a dataset that includes range measurements as well as wheel odometry of a small robot.
While driving through a labyrinth, the distance to fixed points is measured with wireless ultra-wide-band (UWB) modules.
They are able to provide accurate range measurements, but artificial metallic obstacles were added to provoke NLOS effects.
A top-view camera system provides a centimeter-level precise ground truth and allows to evaluate the accuracy of estimation algorithms.
As \autoref{fig:Histogram} shows, the error distribution is asymmetric and right skewed with a mean of \SI{0.12}{\meter} and outliers up to \SI{1}{\meter}.
Since the robot's motion is restricted in a plane, the system's state is simply a 2D-pose.
In difference to earlier work \cite{Pfeifer2017}, we omit the estimation of a common offset for all range measurements.
The maximum-likelihood problem is composed of one range and odometry measurement for each time step.
Further details can be found in \cite{Pfeifer2017}.
\autoref{tab:std_uwb} summarizes the noise properties that are used for the non-robust factors.

\begin{table}[tbph]
	\centering
	\caption{Noise Properties of the UWB Indoor Dataset}
	\label{tab:std_uwb}
	\begin{tabular}{@{}lc@{}l}
		\toprule
		\textbf{Factor} 		& \textbf{Square Root Information} $\sqrtinfo$			\\ \midrule
		Range &
		$\left(\SI{0.1}{\meter}\right)^{-1}$ 								\\\midrule
		Odometry 	&
		$\diag
		\begin{pmatrix}
			\SI{0.01}{\meter\per\second} \\
			\SI{0.01}{\meter\per\second} \\
			\SI{0.01}{\meter\per\second}
		\end{pmatrix}^{-1}$																	\\ \bottomrule
	\end{tabular}
\end{table}

\subsection{GNSS Localization}

\begin{figure}[tbph]
	\centering
	\includegraphics[scale=0.195]{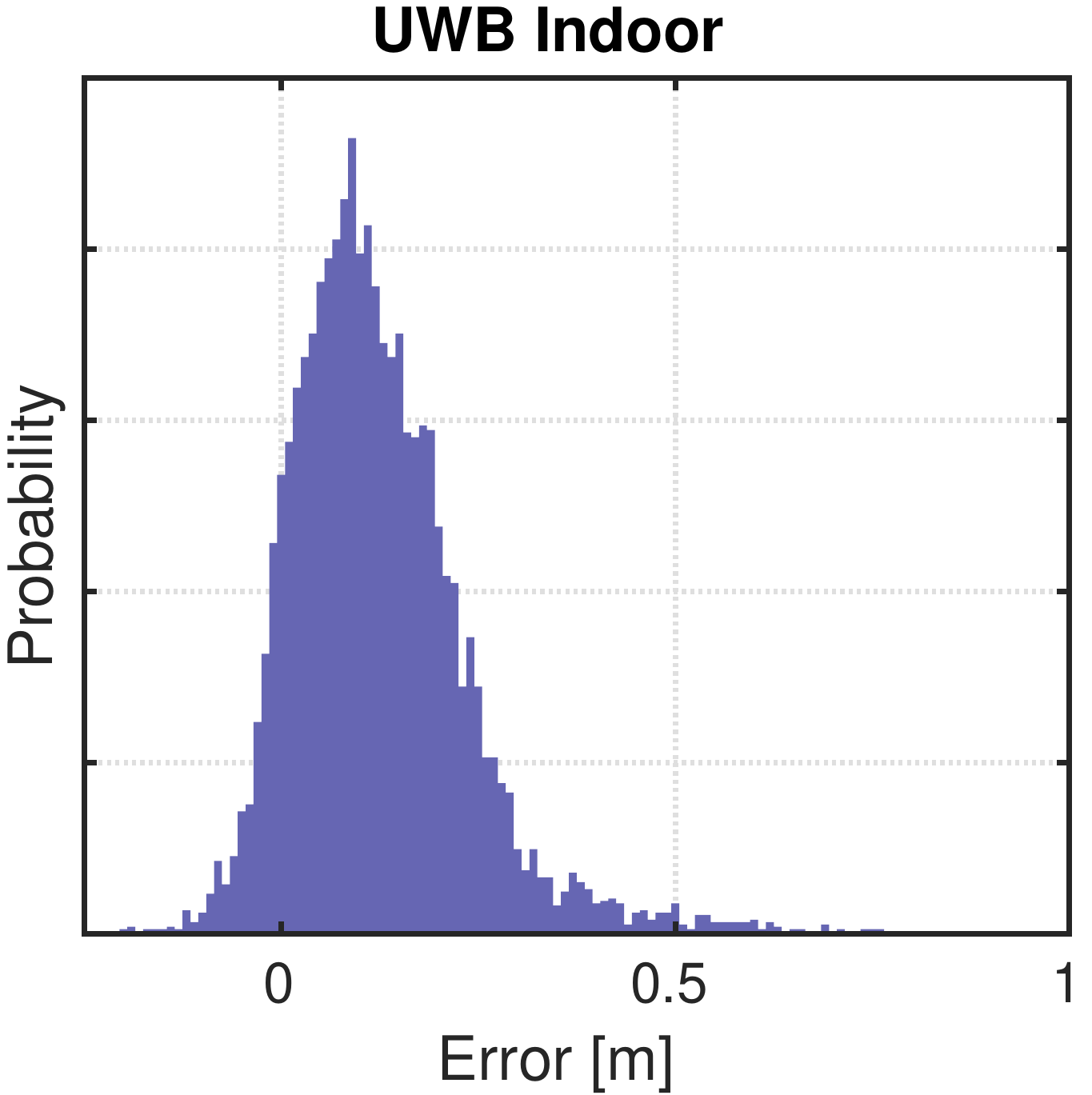}
	\includegraphics[scale=0.195]{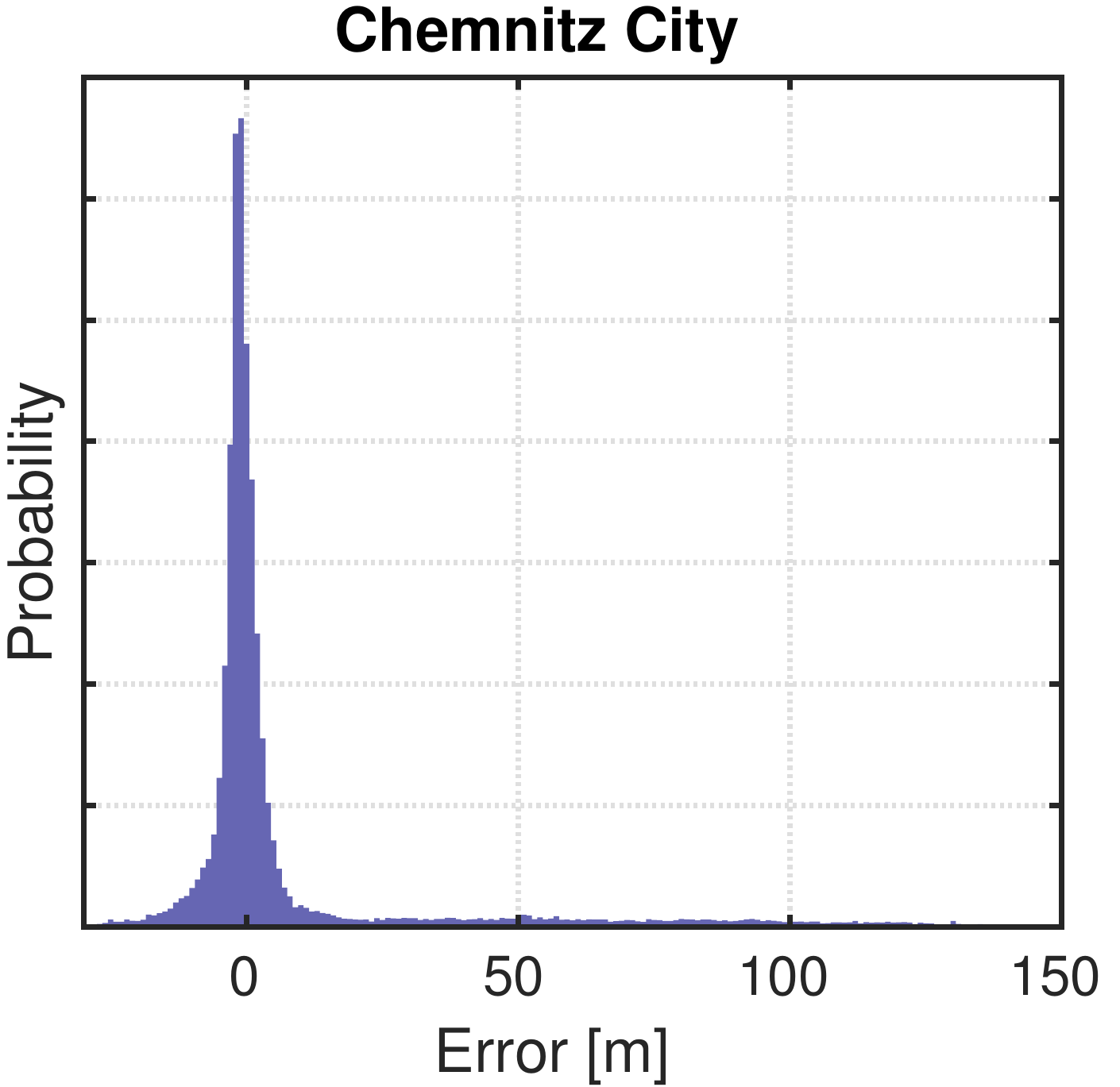}
	\includegraphics[scale=0.195]{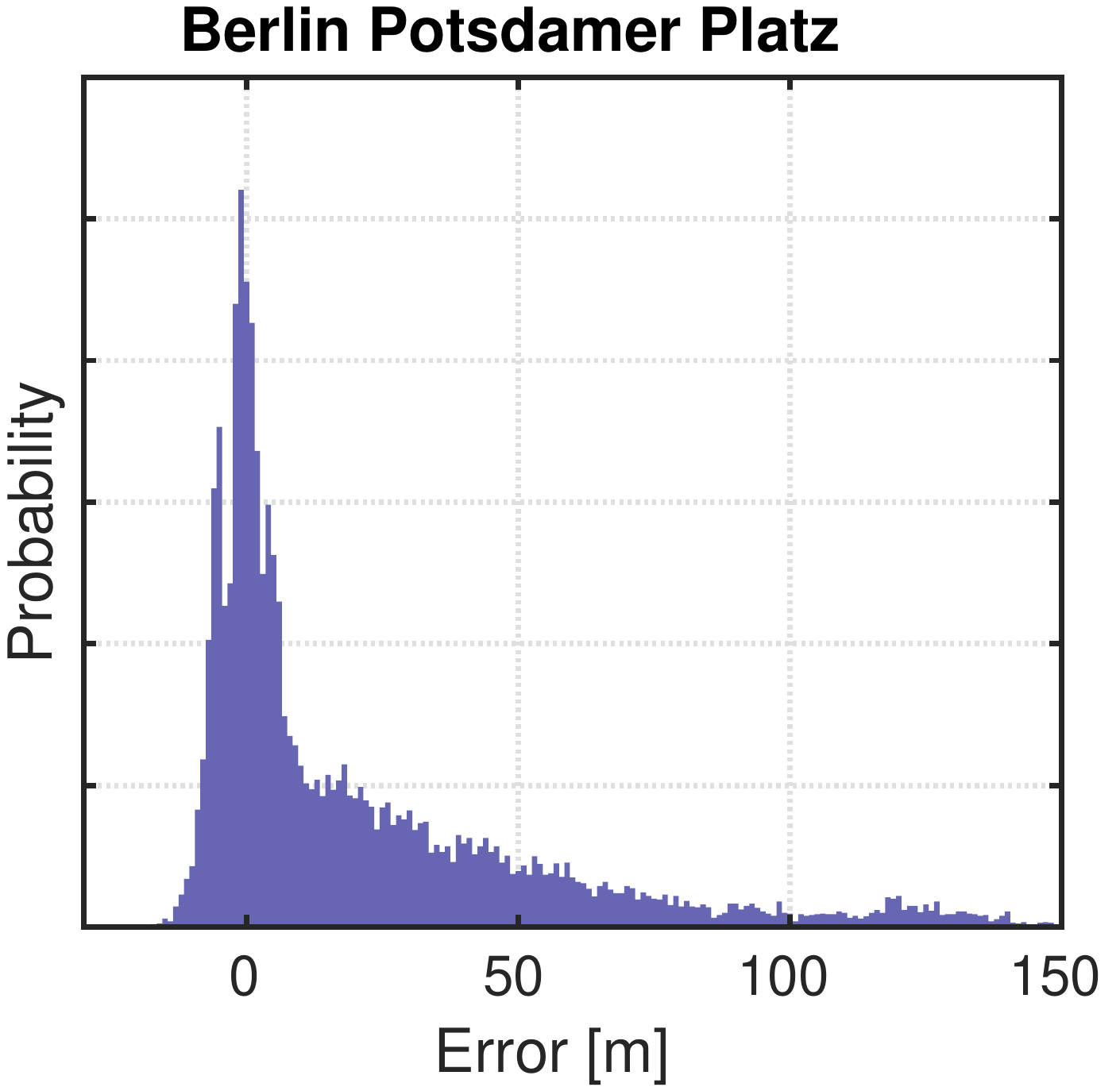}
	\includegraphics[scale=0.195]{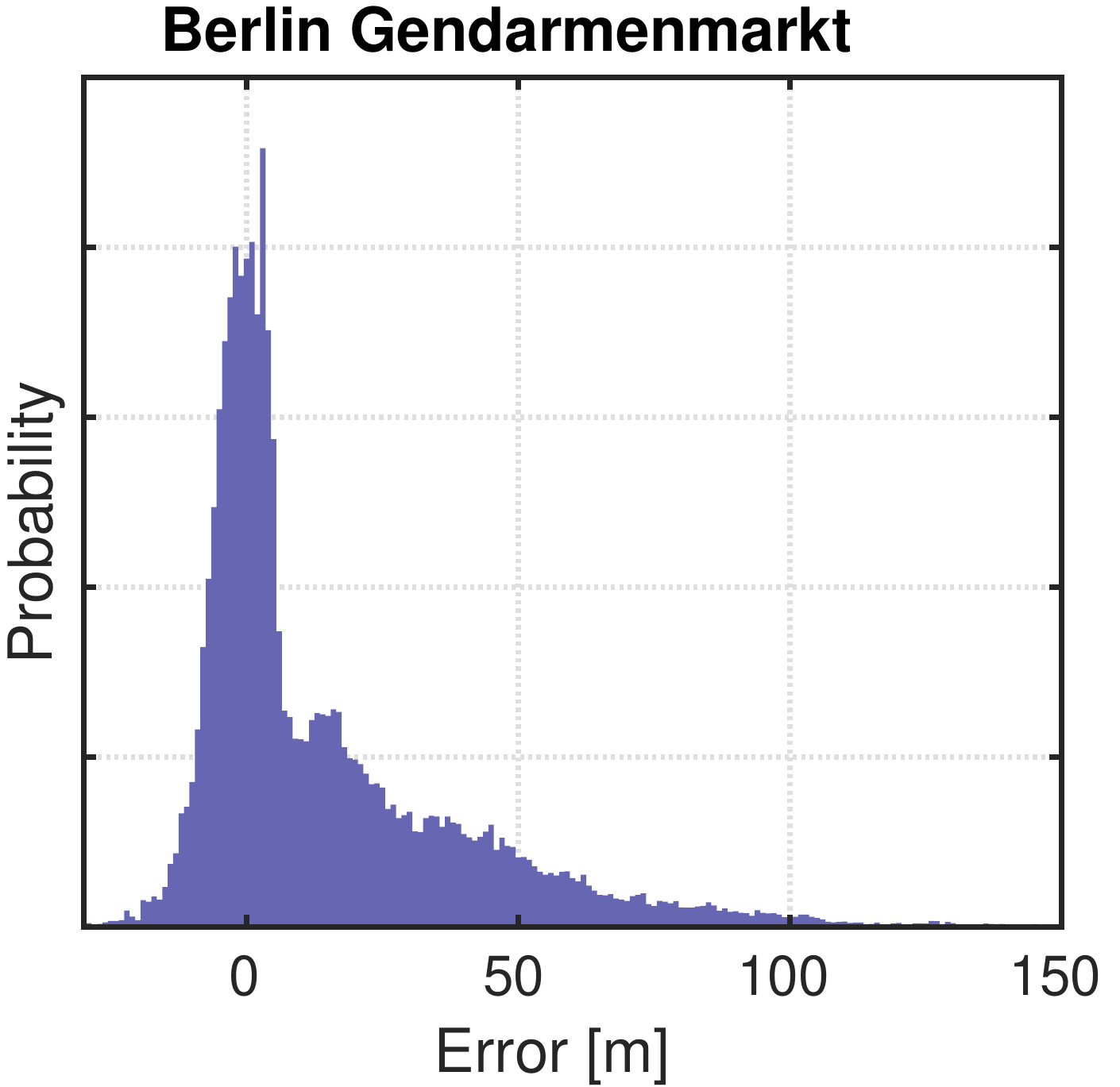}
	\includegraphics[scale=0.195]{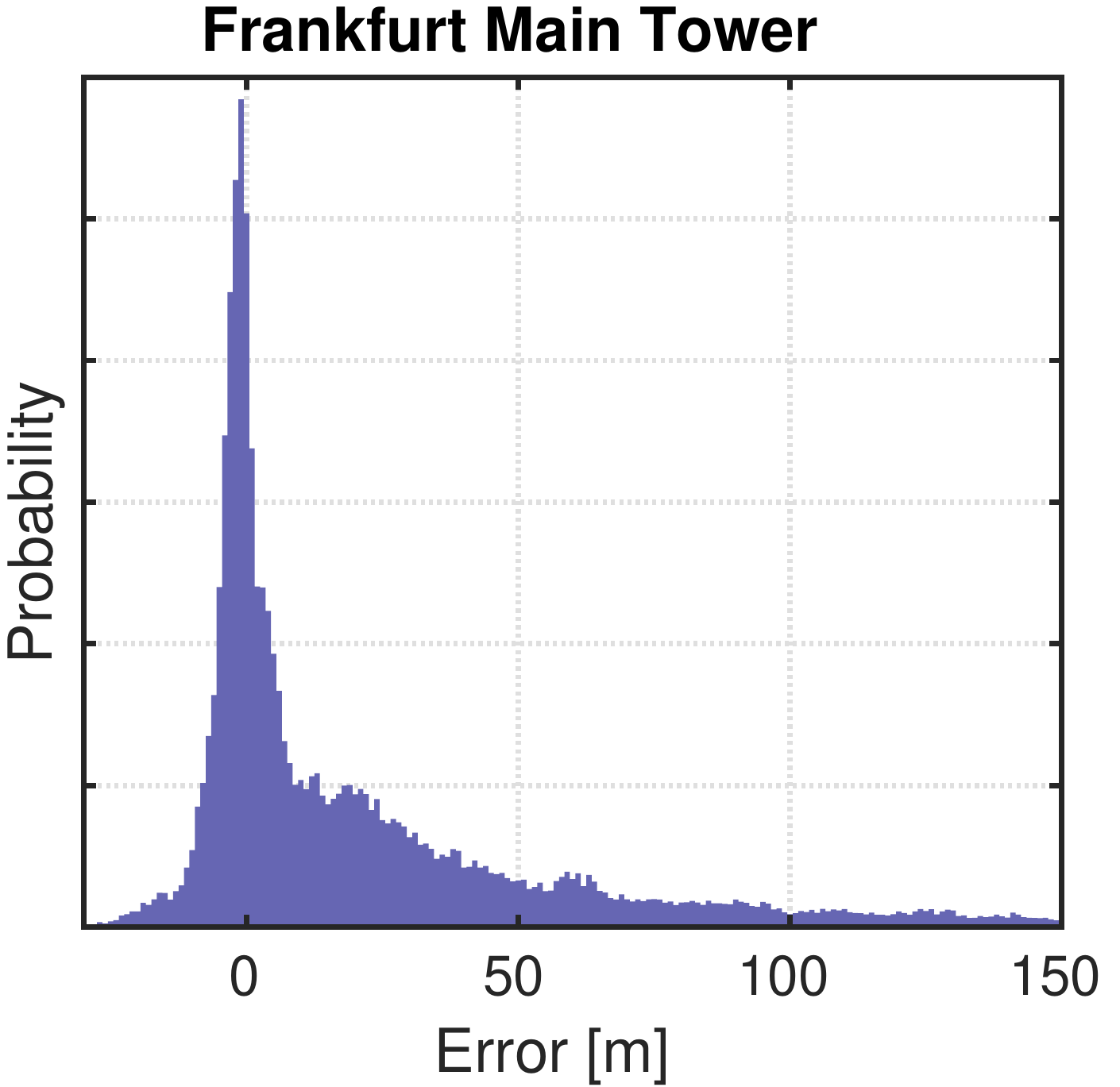}
	\includegraphics[scale=0.195]{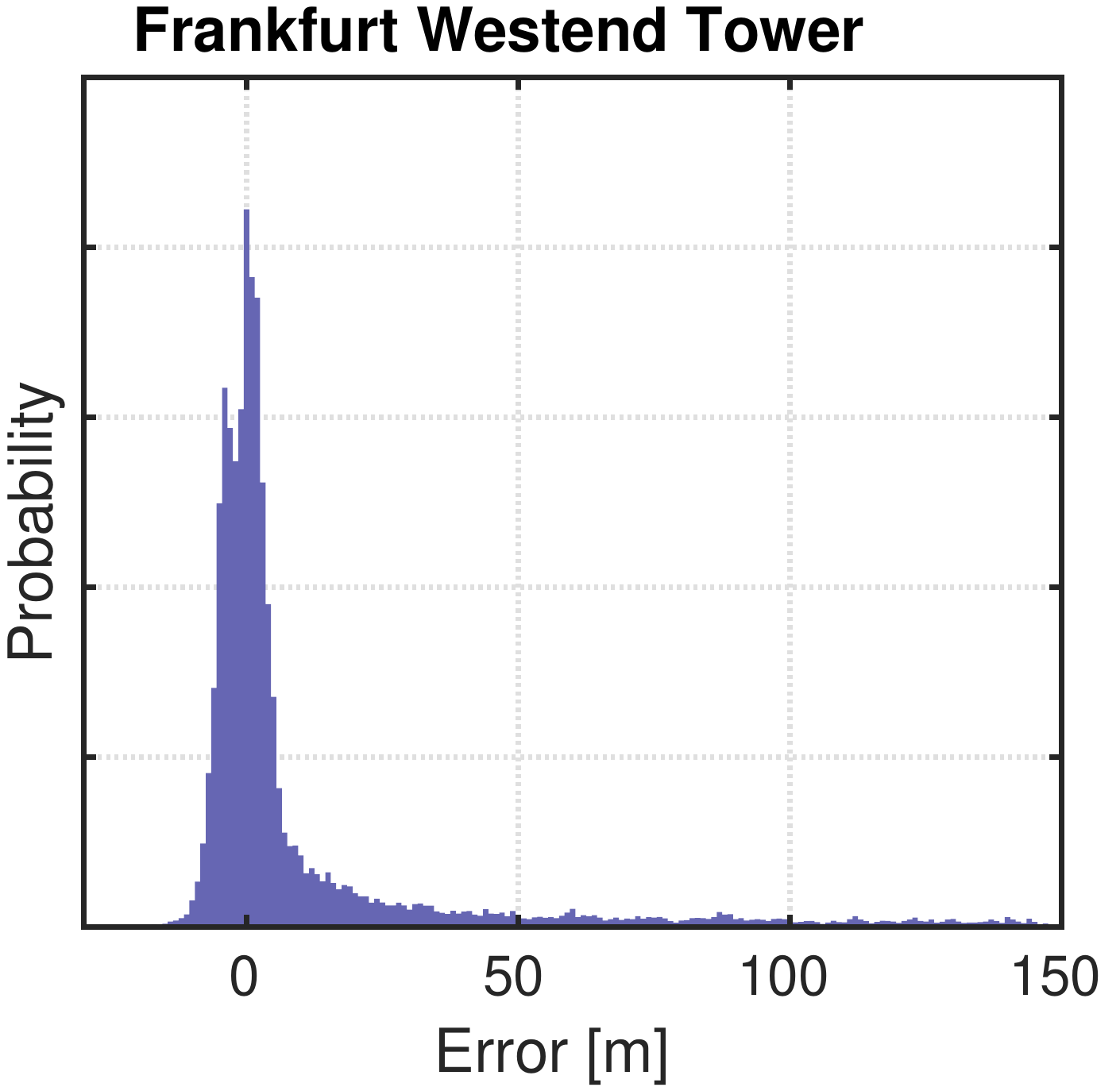}
	\caption{Distributions of the (pseudo) range error of the UWB and GNSS datasets. Please note the differently scaled first x-axis. All data sets show a more or less distinct right skewness, which is caused by NLOS effects.}
	\label{fig:Histogram}
\end{figure}

Besides the robotic example, we evaluate the proposed algorithm on several real world GNSS datasets.
Along with the older Chemnitz City dataset from \cite{Suenderhauf2013a}, we use the four smartLoc datasets \cite{Reisdorf2016} which are recorded in the major cities Frankfurt and Berlin.
Due to the urban environment, they contain a high proportion of NLOS measurements, as their error histogram in \autoref{fig:Histogram} shows.
All datasets combine the wheel odometry of a driving car with a set of pseudorange measurements from a mass-market receiver.
As ground truth, a NovAtel differential GNSS receiver supported by a high-end inertial measurement unit is used.

The vehicle's pose is estimated according to the Cartesian ECEF coordinate system and the rotation around its upright axis.
Since the pseudorange measurements are biased by the receivers drifting clock error, it is estimated along with its derivation.
In difference to the UWB dataset, there are multiple (pseudo) range measurements for each time step. Their number depends on the number of observable satellites.
The dynamic of the clock error is described with a constant clock error drift model (CCED).
Pseudorange, wheel odometry and CCED factor are described in detail in \cite{Pfeifer2018}.
\autoref{tab:std_gnss} summarizes the noise properties of all GNSS datasets.
Since they are recorded with different sensors, the values for odometry and the CCED model differ.
While the pseudorange noise of the Chemnitz City dataset is fixed, the smartLoc datasets include an individual value for each measurement that is provided by the GNSS receiver.

\begin{table}[tbp]
	\centering
	\caption{Noise Properties of the GNSS Datasets}
	\label{tab:std_gnss}
	\setlength\tabcolsep{4pt}
	\begin{tabular}{@{}lcc@{}l}
		\toprule
		\multirow{2}{*}{\textbf{Factor}} 		& \multicolumn{2}{c}{\textbf{Square Root Information $\sqrtinfo$}}						\\ 
					& \textbf{Chemnitz City}	& \textbf{smartLoc}												\\ \midrule
		Pseudorange &
		$\left(\SI{10}{\meter}\right)^{-1}$ &  *							\\\midrule
		Odometry 	&
		$\diag
		\begin{pmatrix}
			\SI{0.05}{\meter\per\second}   \\
			\SI{0.03}{\meter\per\second}   \\
			\SI{0.03}{\meter\per\second}   \\
			\SI{0.006}{\radian\per\second}
		\end{pmatrix}^{-1}$&
		$\diag\begin{pmatrix}
			\SI{0.05}{\meter\per\second}   \\
			\SI{0.03}{\meter\per\second}   \\
			\SI{0.03}{\meter\per\second}   \\
			\SI{0.002}{\radian\per\second}
		\end{pmatrix}^{-1}$																\\ \midrule
		CCED Model & 
		$\diag
		\begin{pmatrix}
		\SI{0.1}{\meter}\\
		\SI{0.009}{\meter\per\second}\\
		\end{pmatrix}^{-1}$&
		$\diag
		\begin{pmatrix}
		\SI{0.05}{\meter}\\
		\SI{0.01}{\meter\per\second}\\
		\end{pmatrix}^{-1}$																	\\ \bottomrule
		\\
		\multicolumn{3}{l}{* individually estimated by the GNSS receiver}\\
	\end{tabular}
\end{table}

\section{Evaluation}\label{sec:results}
\begin{table*}[t]
	\centering
	\caption{Results of the Final Evaluation.}
	\label{tab:result}
	\setlength\tabcolsep{4pt}
	\begin{tabular}{@{}lllllllllllll@{}}
		\toprule
		& \multicolumn{2}{c}{Indoor UWB} & \multicolumn{2}{c}{Chemnitz} & \multicolumn{2}{c}{Berlin PP} & \multicolumn{2}{c}{Berlin GM} & \multicolumn{2}{c}{Frankfurt MT} & \multicolumn{2}{c}{Frankfurt WT} \\
		\multirow{-2}{*}{Algorithm} & ATE [m] & Time [s] & ATE [m] & Time [s] & ATE [m] & Time [s] & ATE [m] & Time [s] & ATE [m] & Time [s] & ATE [m] & Time [s] \\ \midrule
		Gaussian &  0.1212 & 26.0 & 30.0 & 56.7 & 29.2 & 9.84 & 13.38 & 47.3 & 30.97 & 54.8 & 23.54 & 32.2 \\
		DCS &  0.1443 & 55.9 & 4.403 & 52.6 & 25.04 & 15.7 & 19.11 & 62.3 & \bestResult{13.25} & 65.0 & 11.39 & 35.8 \\
		cDCE &  0.1210 & 53.7 & 4.326 & 52.3 & 17.91 & 15.2 & 14.59 & 64.6 & 14.93 & 68.8 & 11.12 & 35.5 \\
		Static MM &  0.1569 & 24.6 & 4.127 & 55.3 & 27.25 & 15.2 & 15.73 & 63.1 & 18.1 & 64.4 & 9.467 & 37.9 \\
		Static SM &  0.1185 & 39.3 & 4.463 & 55.1 & 18.36 & 19.8 & \bestResult{12.52} & 63.8 & 20.15 & 69.6 & 10.82 & 37.2 \\
		Adaptive MM &  0.0666 & 41.0 & 2.562 & 88.4 & 26.65 & 28.0 & 16.8 & 124.0 & 15.39 & 111.0 & 6.995 & 67.9 \\
		Adaptive SM &  \bestResult{0.0651} & 46.4 & \bestResult{2.559} & 81.0 & \bestResult{12.96} & 24.2 & 15.67 & 110.0 & 13.89 & 96.9 & \bestResult{6.669} & 58.6  \\
		\bottomrule
	\end{tabular}
\end{table*}
This section gives an overview over the performance of the proposed adaptive mixture approach.
Beside the estimation accuracy, we want to demonstrate the good convergence properties of the adapted error model.
We compare our approach against static mixture models as well as the state-of-the-art algorithm DCS \cite{Agarwal2013} and cDCE \cite{Pfeifer2017}.
We use the absolute trajectory error (ATE) in the XY-plane as error metric, since it can be applied to both estimation problems.
The source code as well as the datasets are online available as part of our robust sensor fusion C++ library libRSF\footnote{\url{https://mytuc.org/libRSF}}.

\subsection{Implementation Details}

The least squares optimization of the state estimation problem is implemented with the Ceres solver \cite{Agarwal}.
To keep the runtime bounded, we use a simplistic sliding window approach that removes factors older than $t_{SW} = \SI{60}{\second}$.
Even when we use recorded datasets, we process the data as it were in real time.
Hence, the estimation problem is solved in every time step without using future measurements.

\subsection{Parametrization of the GMM}

The initial parametrization $\Params^{GMM}_{t=0}$ of the GMM can be determined in different ways.
It can be estimated with a simple clustering approach like k-means or defined with prior knowledge about the estimation problem.
We assume that the (pseudo) range measurements are corrupted by outliers that have a larger spread than the valid measurements.
Therefore, we define the initial GMM with equal-weighted, zero-mean components that differ only in their square root information.
As defined with \autoref{eqn:init_gmm} for a n-component GMM, we scale each matrix $\sqrtinfoj$ with factor $10^{1-j}$ based on the square root information for the non-robust Gaussian case $\sqrtinfo$.
For the smartLoc datasets, we also use a value of $\sqrtinfo = (\SI{10}{\meter})^{-1}$ to be consistent with the Chemnitz City dataset.
The algorithms with a static GMM rely on the same values.
\begin{equation}
\left.
\begin{matrix}
\sqrtinfoj &=& \sqrtinfo \cdot 10^{1-j}\\
\mean_j &=& 0\\
\weight_j &=& \frac{1}{n}
\end{matrix}
\right\rbrace 
\forall j = 1 \dots n
\label{eqn:init_gmm}
\end{equation}

\begin{figure}[tbph]
	\centering
	\includegraphics[scale=0.2]{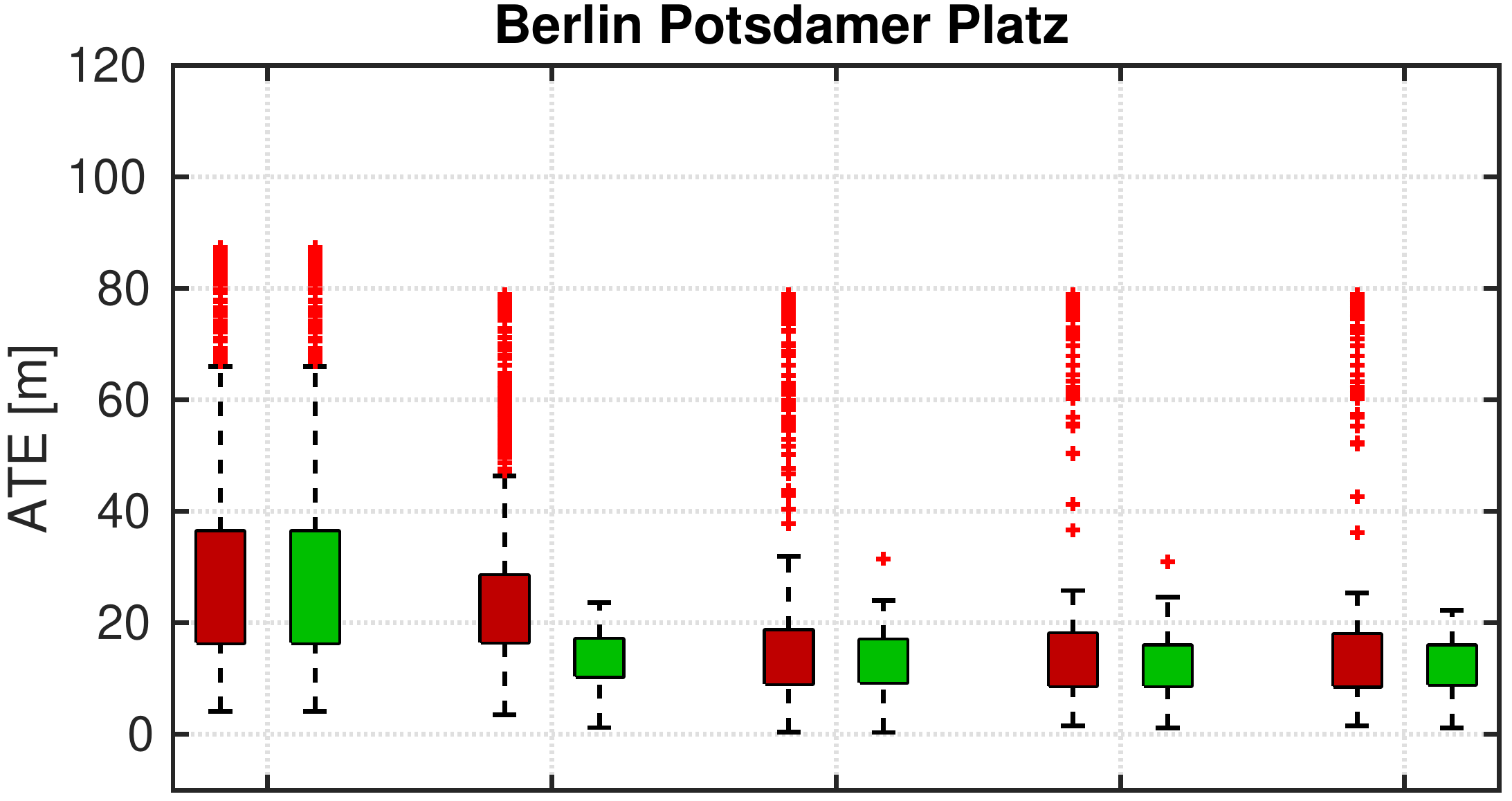}
	\includegraphics[scale=0.2]{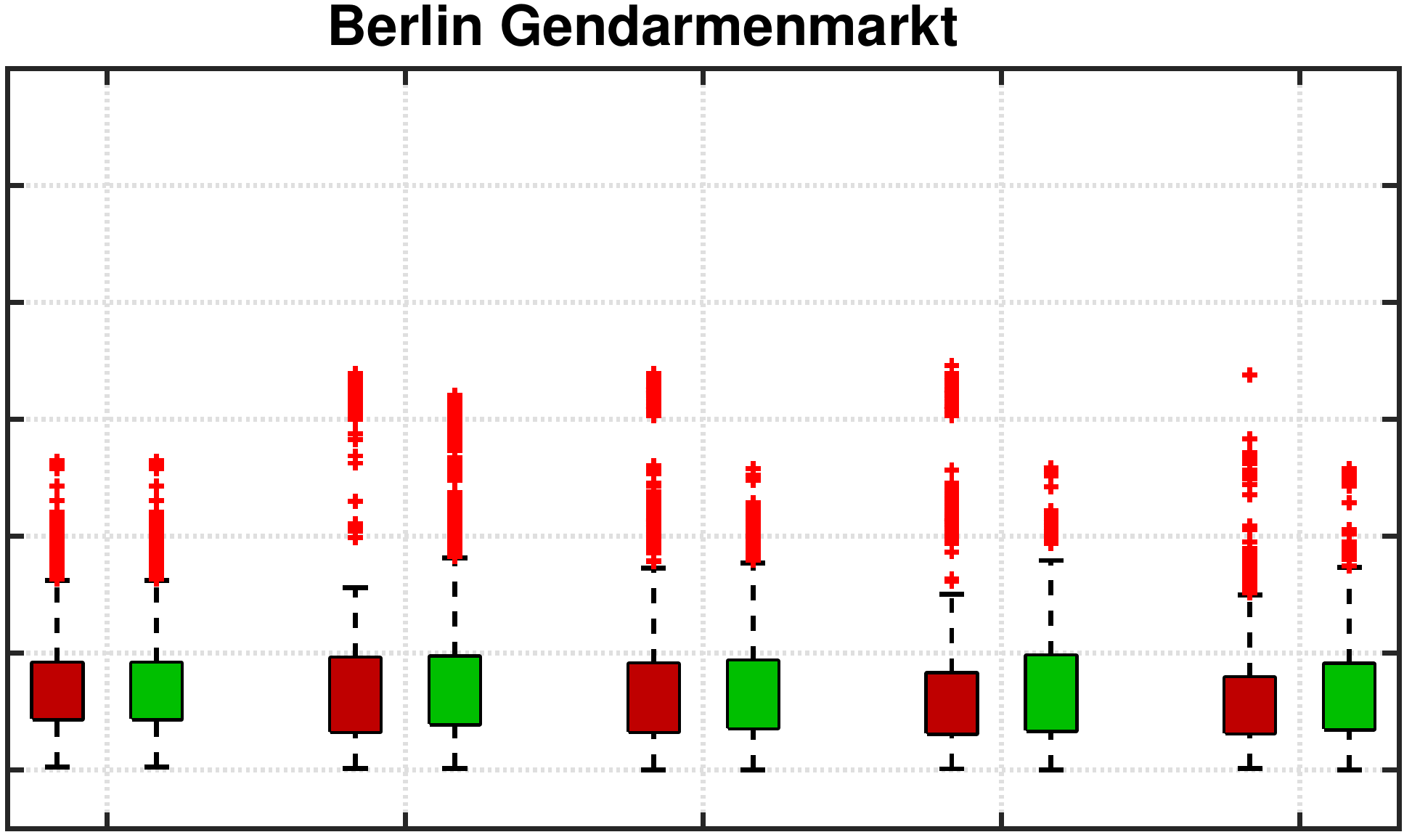}
	\includegraphics[scale=0.2]{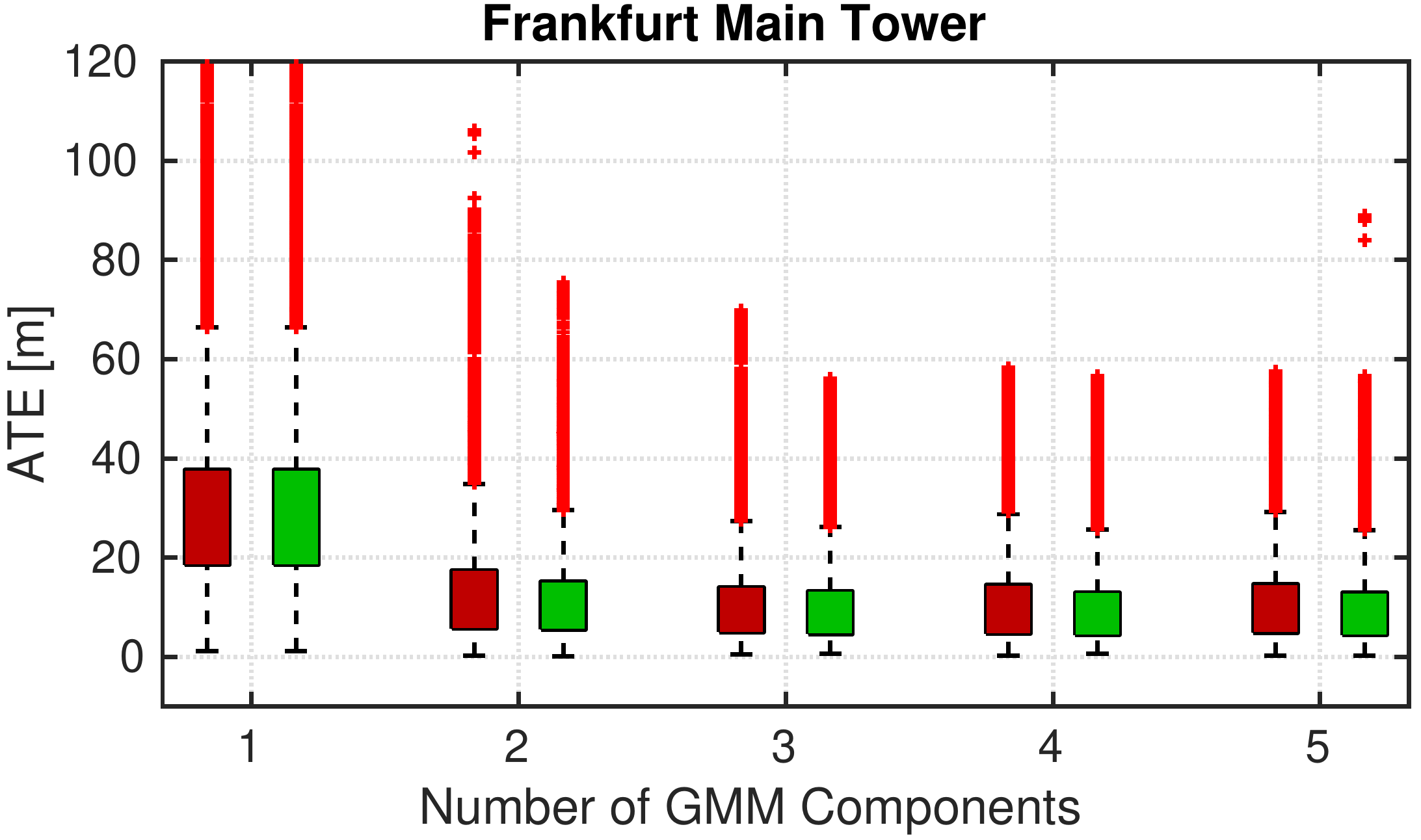}
	\includegraphics[scale=0.2]{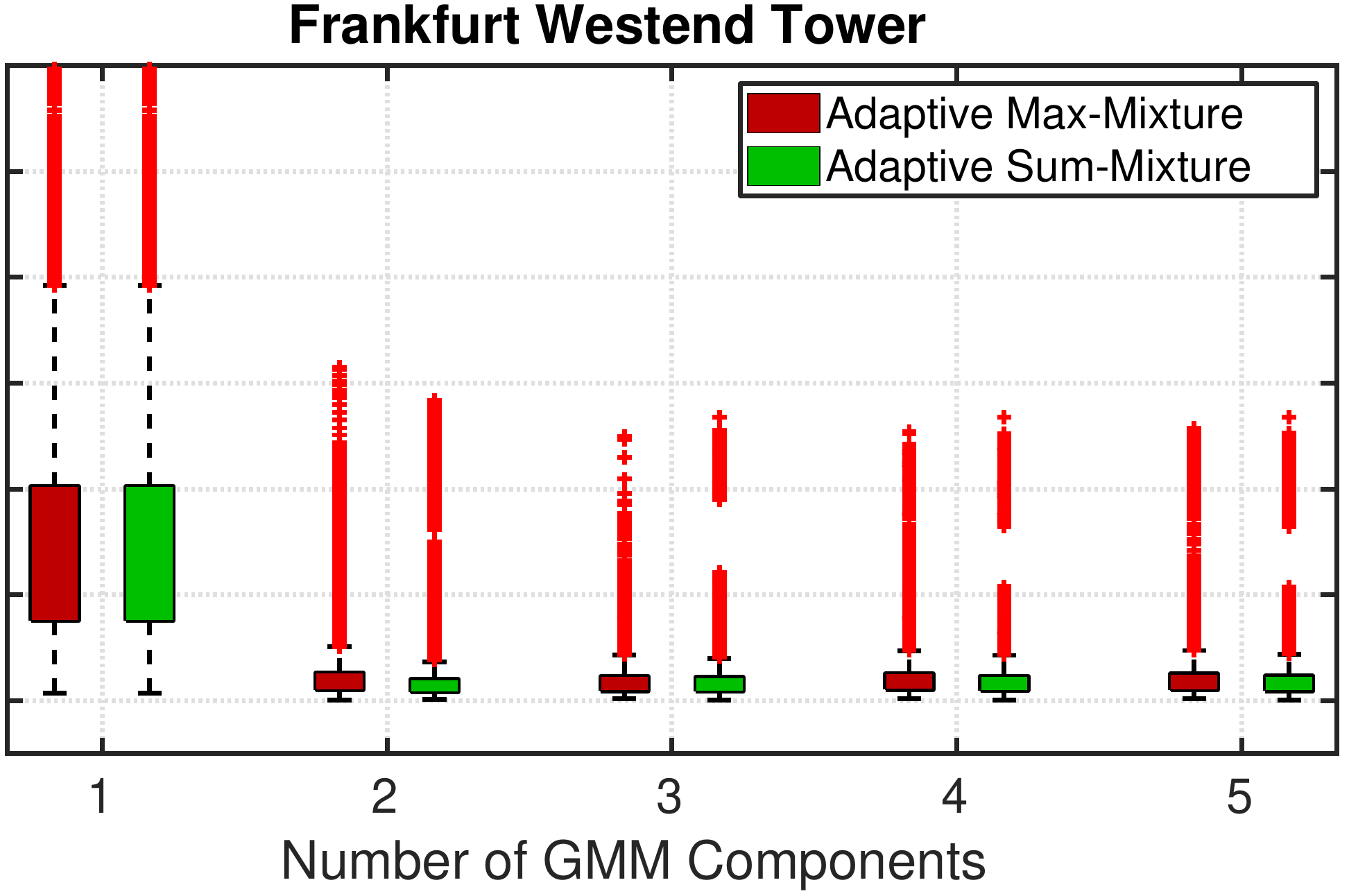}
	\caption{Position error for the smartLoc GNSS datasets depending on the number of components of the GMM. Using more than two components has no advantage regarding the error. Please note the shared axes.}
	\label{fig:CompEval}
\end{figure}

For the automatic selection of the number of GMM components, we examined classical static criteria such as BIC \cite{Schwarz1978} and AIC \cite{Akaike1974}.
Since the sensor data is strongly multimodal, both lead to higher numbers like $n=5, 6$.
Regarding the trajectory error, our evaluation in \autoref{fig:CompEval} has shown no advantages for higher numbers, we set $n=2$ for all datasets.
\begin{figure}[tbph]
\centering
\includegraphics[scale=0.2]{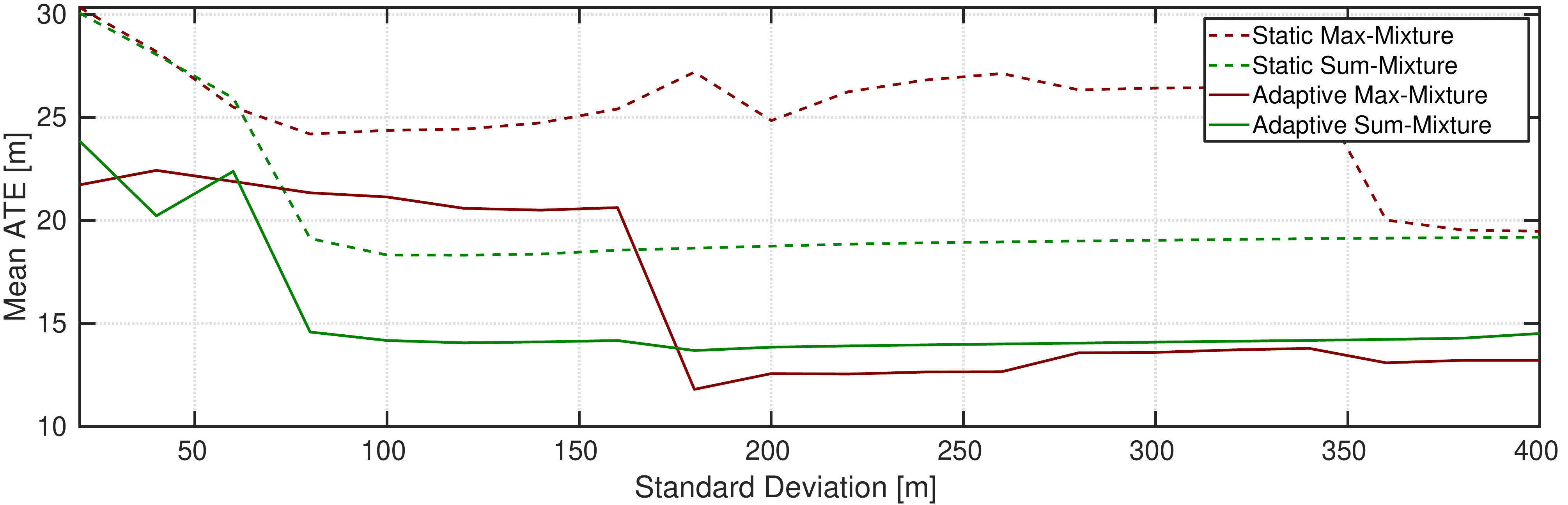}
\caption{Mean position error for the dataset ``Berlin Potsdamer Platz'' depending on the initialization of the second GMM component's standard deviation. The exact sum-mixture approach converges for a wider range of values, but both adaptive approaches outperform their static variants.}
\label{fig:ParamEval}
\end{figure}
Although the proposed parametrization is applied in our final evaluation, we want to demonstrate that the self-tuning algorithms are robust regarding their initialization.
Therefore, we evaluated the mean ATE of adaptive MM and SM for different standard deviations of the second GMM component.
\autoref{fig:ParamEval} shows that there is a wide basin of convergence for both algorithms, although Sum-Mixture converges slightly better.
Both algorithms surpass their static variants significantly for any parametrization.

\subsection{Position Accuracy}

\autoref{tab:result} summarizes the mean ATE as well as the runtime of all algorithms.
The smallest trajectory error is marked in green.
For 5 out of 6 datasets, adaptive mixture models are able to reduce the impact of erroneous measurements over static ones.
They are the only ones that reduce the ATE of the indoor dataset significantly.
This can be explained by the shifted distribution, which cannot be compensated by other robust methods.
The proposed adaptive SM algorithm achieved the best results in 4 cases.
DCS achieved a slightly better result on the ``Frankfurt WT'' dataset, which has a relatively low number of NLOS measurements.
However, it performs significantly worse on other datasets.
A special case is the ``Berlin GM'' dataset, where only the static SM algorithm were able to reduce the ATE.
We have to investigate further which characteristic caused this result.
The adaptive MM algorithm achieves comparable but slightly worse results than the SM variant.
Since there is no significant difference in runtime, we would prefer the exact adaptive SM algorithm.

\subsection{Runtime}

The computation times in \autoref{tab:result} are measured on an Intel i7-7700 system.
As a rule of thumb, the adaptive algorithms require twice as much time as the other robust ones.
Nevertheless, both proposed algorithms are significantly faster than the recording time of the datasets.

\section{Conclusion}\label{sec:conclusion}
In this paper, we introduced a novel approach to combine multimodal and self-tuning error models to improve least squares optimization.
Based on the EM-algorithm, we are able to efficiently adapt a Gaussian mixture model during the state estimation process.
Therefore, the burden of parametrization could be relaxed for Max-Mixture \cite{Olson2012} and Sum-Mixture \cite{Rosen2013} based algorithms.
We compared our work to a set of state-of-the-art algorithms on real world datasets and showed their improved performance.
Especially for the adaptive Sum-Mixture algorithm, the well-behaved convergence could be demonstrated experimentally.

Still open for future research is the question after the optimal number of Gaussian components.
Although ``two'' is a sufficient compromise between flexibility and runtime, this does not necessarily apply to scenarios beside GNSS.
Interesting could also be the transfer to SLAM problems that are defined by relative rather than absolute measurements.

\appendix

\subsection{Expectation-Maximization for GMMs}\label{Appendix_EM}

The expectation step is defined by:
\begin{equation}
\alpha_{ij } = \frac{\prob(\errori|\weight_j, \mean_j, \sqrtinfoj)}{\sum_{k=1}^{n}\prob(\errori|\weight_k, \mean_k, \sqrtinfok)}
\label{eqn:e-step}
\end{equation}
\begin{equation*}
\begin{split}
&\text{ with } \prob(\errori|\weight, \mean, \sqrtinfo) = \\
&\weight \cdot \det\left( \sqrtinfo \right) \cdot \exp \left( -\frac{1}{2} \left\| \sqrtinfo \left( \errori - \mean \right) \right\| ^2 \right)
\end{split}
\end{equation*}

The maximization step is defined by:
\begin{equation}
\begin{split}
\weight_j = \frac{1}{m} \sum_{i=1}^{m} \alpha_{ij} \hspace{0.7cm}
\mean_j = \frac{\sum_{i=1}^{m} \alpha_{ij} \errori}{\sum_{i=1}^{m} \alpha_{ij}}\\
{\sqrtinfoj}\tran \cdot \sqrtinfoj = \left( \frac{\sum_{i=1}^{m} \alpha_{ij} \left( \errori - \mean_j \right)\left( \errori - \mean_j \right)^{T} }{\sum_{i=1}^{m} \alpha_{ij}} \right) ^{-1}
\end{split}
\label{eqn:m-step}
\end{equation}

\bibliographystyle{IEEEtranN}
\bibliography{IEEEabrv,CleanBib}

\end{document}

%% file: math.tex
\DeclareMathOperator*{\argmin}{\arg\!\min}
\DeclareMathOperator*{\argmax}{\arg\!\max}
\DeclareMathOperator{\diag}{\operatorname{diag}}

\newcommand{\vect}[1]{\mathbf{ #1}}
\newcommand{\est}[1]{\hat{#1}}
\newcommand{\opt}[1]{#1^\ast}
\newcommand{\tran}{^\mathsf{T}}

\newcommand{\State}{\vect{X}}
\newcommand{\StateOpt}{\opt{\vect{X}}}
\newcommand{\StateEst}{\vect{\est{X}}}

\newcommand{\state}{\vect{x}}
\newcommand{\stateOf}[2]{\state^{#1}_{#2}}
\newcommand{\statei}{\stateOf{}{i}}
\newcommand{\stateiEst}{\vect{\est{x}}_{i}}

\newcommand{\Measurement}{\vect{Z}}

\newcommand{\measurement}{\vect{z}}
\newcommand{\measurementOf}[2]{\measurement^{#1}_{#2}}
\newcommand{\measurementi}{\measurementOf{}{i}}

\newcommand{\error}{\vect{e}}
\newcommand{\errorOf}[2]{\error^{#1}_{#2}}
\newcommand{\errori}{\errorOf{}{i}}
\newcommand{\erroriEst}{\vect{\est{e}}_{i}}

\newcommand{\prob}{\vect{P}}
\newcommand{\probOf}[2]{\vect{P}\left( #1 | #2\right) }

\newcommand{\cov}{\boldsymbol{\Sigma}}
\newcommand{\mean}{{\boldsymbol{\mu}}}
\newcommand{\weight}{w}
\newcommand{\sqrtinfo}{\boldsymbol{\mathcal{I}}^{\frac{1}{2}}}
\newcommand{\sqrtinfoOf}[1]{\boldsymbol{\mathcal{I}}_{#1}^{\frac{1}{2}}}
\newcommand{\sqrtinfoj}{\sqrtinfoOf{j}}
\newcommand{\sqrtinfok}{\sqrtinfoOf{k}}

\newcommand{\Params}{\boldsymbol{\theta}}
\newcommand{\ParamsEst}{\boldsymbol{\est{\theta}}}
\newcommand{\observedvar}{\vect{O}}
\newcommand{\hiddenvar}{\vect{H}}